\documentclass{article}

\usepackage[final]{corl_2018} 

\usepackage{amsfonts}
\usepackage{algorithm}
\usepackage{algpseudocode}
\usepackage{graphicx}
\usepackage{subcaption}
\usepackage{amsmath}
\usepackage{boldline}

\title{Energy-Based Hindsight Experience Prioritization}

\author{Rui Zhao {\normalfont and} Volker Tresp\\
Ludwig Maximilian University, Munich, Germany\\  
Siemens AG, Corporate Technology, Munich, Germany\\
\texttt{\{ruizhao,volker.tresp\}@siemens.com}\\
}

\begin{document}
\maketitle







\begin{abstract}
In Hindsight Experience Replay (HER), a reinforcement learning agent is trained by treating whatever it has achieved as virtual goals.
However, in previous work, the experience was replayed at random, without considering which episode might be the most valuable for learning.
In this paper, we develop an energy-based framework for prioritizing hindsight experience in robotic manipulation tasks.
Our approach is inspired by the work-energy principle in physics.
We define a trajectory energy function as the sum of the transition energy of the target object over the trajectory. 
We hypothesize that replaying episodes that have high trajectory energy is more effective for reinforcement learning in robotics.
To verify our hypothesis, we designed a framework for hindsight experience prioritization based on the trajectory energy of goal states.
The trajectory energy function takes the potential, kinetic, and rotational energy into consideration.
We evaluate our Energy-Based Prioritization (EBP) approach on four challenging robotic manipulation tasks in simulation.
Our empirical results show that our proposed method surpasses state-of-the-art approaches
in terms of both performance and sample-efficiency on all four tasks, without increasing computational time.
A video showing experimental results is available at \url{https://youtu.be/jtsF2tTeUGQ}.

\end{abstract}



\section{Introduction}

Reinforcement learning techniques \cite{sutton1998reinforcement} combined with deep neural networks \cite{goodfellow2016deep, zhao2017two} led to great successes in various domains, such as playing video games \cite{mnih2015human}, challenging the World Go Champion \cite{silver2016mastering}, conducting goal-oriented dialogues \cite{bordes2016learning, zhao2018improving, zhao2018learning, zhao2018efficient}, and learning autonomously to accomplish robotic tasks \cite{ng2006autonomous, peters2008reinforcement, levine2016end, chebotar2017path,andrychowicz2017hindsight}.

In robotic tasks, autonomous agents are expected to achieve multiple goals in different scenarios. 
Standard approaches are based on goal-conditioned policies that allow agents to learn different polices concurrently \cite{schaul2015universal, schmidhuber2013powerplay, deisenroth2014multi, zhu2017target, held2017automatic}. 
Alternatively, the agent exploits what alternative goals it has achieved, learns from these achieved states, and further attempts to achieve the real goal. 
This kind of goal-conditioned curriculum learning has recently been introduced as hindsight experience replay (HER) \cite{andrychowicz2017hindsight}. 
HER lets an agent learn from undesired outcomes and tackles the problem of sparse rewards in Reinforcement Learning (RL). 

However, HER samples episodes uniformly from the memory buffer for replay. 
Subsequently, in the selected episode, the virtual goals are sampled randomly at a future timestep with respect to a randomly chosen state.
The replay process does not consider which episodes or states might be the most valuable for learning \cite{plappert2018multi}.
It would be more efficient in training to prioritize the more important and valuable episodes. 
The challenge now is how to judge which episodes are more valuable for learning. 
We define the trajectory energy as the sum of the transition energy of the object over all timesteps of the trajectory, see more detail in Section \ref{sec:trajectory_energy_function}.
Our hypothesis is that the trajectory energy is an effective metric for indicating which episode is more difficult to achieve. 
This is readily true in most robotic manipulation tasks. 
Imagine a robot arm with an end effector, attempting to pick up an object on the floor and place it on a shelf. 
The achieved goal state is the position of the object at each timestep. 
In an unsuccessful scenario, the robot arm attempts to reach the object but fails, leaving the object on the floor. 
The total energy of the object, including the potential energy and the kinetic energy, does not change during the episode because of Newton's laws of motion or, equivalently, the work-energy principle. 
Therefore, the trajectory energy of the object remains zero. 
In a somewhat better scenario, the robot arm picks up the object successfully, but accidentally drops the object before it reaches the shelf. 
In this case, the trajectory energy of the object rises because the robot arm does work on the object. 
This case can be explained by the work-energy principle in physics. 
In a successful scenario, the robot arm picks up the object and transfers the object to the goal position on the shelf. 
Here, the trajectory energy is the highest among all three scenarios; the robot arm does the most work on the object and fulfills the task. 
Obviously, the successful episode is the most valuable for replay. 
The first scenario is the least important episode because the object state is barely correlated with the trajectory of the robot arm. 
In this robotic task example, we can see that the trajectory energy indeed indicates the importance of episodes.

In almost all robotic tasks, goal states can be expressed with physics quantities. In these cases, the energy-based prioritization method is applicable. Based on the position and the orientation of the object, we can calculate the linear and the angular velocity, as well as the kinetic energy and the rotational energy. Based on the height of the object, we can calculate the potential energy. We callculate the energy increases from state to state as the transiton energy, see Equation (\ref{eq:transition}).  We sum the transition energy over time to have the trajectory energy, see Equation (\ref{eq:trajectory}). Using the trajectory energy function, in our approach, we prioritize the episodes with higher trajectory energy to speed up the training. In order to verify our hypothesis, we implemented and tested our prioritization framework in four simulated robotic tasks. These tasks include pick-and-place with a robot arm and manipulating a block, an egg, and a pen with a Dexterous robot hand. The tasks are simulated with the MuJoCo physics engine \cite{todorov2012mujoco} and runs in the OpenAI Gym environment \cite{brockman2016openai,plappert2018multi}.  

In this paper we propose to use the trajectory energy function of achieved goal states as a metric to evaluate which episodes are more difficult to achieve. Subsequently, we introduce Energy-Based Prioritization (EBP) with hindsight experience replay. EBP prioritizes the trajectories with higher energy during training to improve sample-efficiency. The proposed technique is applicable to any robotic manipulation task, whenever multi-goal off-policy RL algorithms apply. The core idea of EBP is to prioritize the explored episode, which is relatively difficult to achieve. This can be considered as a form of curriculum learning, which selects difficult yet achievable episodes for replay.


\section{Background}
\label{sec:background}

In this section, we introduce the preliminaries, such as the used reinforcement learning approaches and the work-energy principle in physics.    

\subsection{Markov Decision Process}

We consider an agent interacting with an environment. We assume the environment is fully observable, including a set of state $\mathcal{S}$, a set of action $\mathcal{A}$, a distribution of initial states $p(s_0)$, transition probabilities $p(s_{t+1}|s_t, a_t)$, a reward function $r$: $\mathcal{S} \times \mathcal{A} \rightarrow \mathbb{R}$, and also a discount factor $\gamma \in [0,1]$. These components formulate a Markov decision process represented as a tuple, $(\mathcal{S}, \mathcal{A}, p, r, \gamma)$. A policy $\pi$ maps a state to an action, $\pi:\mathcal{S} \rightarrow \mathcal{A}$.

At the beginning of each episode, an initial state $s_0$ is sampled from the distribution $p(s_0)$. Then, at each timestep $t$, an agent performs an action $a_t$ at the current state $s_t$, which follows a policy $a_t=\pi(s_t)$. Afterwards, a reward $r_t=r(s_t, a_t)$ is produced by the environment and the next state $s_{t+1}$ is sampled from the distribution $p(\cdot|s_t, a_t)$. The reward might be discounted by a factor $\gamma$ at each timestep. The goal of the agent is to maximize the accumulated reward, i.e.\ the return, $R_t=\sum_{i=t}^{\infty}\gamma^{i-t}r_i$, over all episodes, which is equivalent to maximizing the expected return, $\mathbb{E}_{s_0}[R_0|s_0]$.

\subsection{Deep Deterministic Policy Gradient}

The objective $\mathbb{E}_{s_0}[R_0|s_0]$ can be maximized using temporal difference learning, policy gradients, or the combination of both, i.e.\ the actor-critic methods \cite{sutton1998reinforcement}. For continuous control tasks, Deep Deterministic Policy Gradient (DDPG) shows promising performance, which is essentially an off-policy actor-critic method \cite{lillicrap2015continuous}. DDPG has an actor network, $\pi: \mathcal{S} \rightarrow \mathcal{A}$, that learns the policy directly. It also has a critic network, $Q: \mathcal{S} \times \mathcal{A} \rightarrow \mathbb{R}$, that learns the action-value function, i.e.\ Q-function $Q^{\pi}$. During training, the actor network uses a behavior policy to explore the environment, which is the target policy plus some noise, $\pi_b = \pi(s) + \mathcal{N}(0,1)$. The critic is trained using temporal difference learning with the actions produced by the actor: $y_t=r_t+\gamma Q(s_{t+1}, \pi(s_{t+1}))$. The actor is trained using policy gradients by descending on the gradients of the loss function, $\mathcal{L}_a=-\mathbb{E}_s[Q(s,\pi(s))]$, where $s$ is sampled from the replay buffer. 
For stability reasons, the target $y_t$ for the actor is usually calculated using a separate network, i.e.\ an averaged version of the previous Q-function networks \cite{mnih2015human,lillicrap2015continuous,polyak1992acceleration}.
The parameters of the actor and critic are updated using backpropagation. 

\subsection{Hindsight Experience Replay}

For multi-goal continuous control tasks, DDPG can be extended with Universal Value Function Approximators (UVFA) \cite{schaul2015universal}. UVFA essentially generalizes the Q-function to multiple goal states $g\in \mathcal{G}$. 
For the critic network, the Q-value depends not only on the state-action pairs, but also depends on the goals: $Q^{\pi}(s_t, a_t, g)=\mathbb{E}[R_t|s_t, a_t, g]$.

For robotic tasks, if the goal is challenging and the reward is sparse, then the agent could perform badly for a long time before learning anything. Hindsight Experience Replay (HER) encourages the agent to learn something instead of nothing. During exploration, the agent samples some trajectories conditioned on the real goal $g$. 
The main idea of HER is that during replay, the selected transitions are substituted with achieved goals $g'$ instead of the real goals. In this way, the agent could get a sufficient amount of reward signal to begin learning. \citet{andrychowicz2017hindsight} show that HER makes training possible in challenging robotic environments. However, the episodes are uniformly sampled in the replay buffer, and subsequently, the virtual goals are sampled from the episodes. More sophisticated replay strategies are requested for improving sample-efficiency \cite{plappert2018multi}.

\subsection{Work-Energy Principle}

Prior to the energy-based hindsight experience prioritization method, we illustrate here the work-energy principle using robotic manipulation examples. 
In physics, a force is said to do work if, when acting, there is a displacement of the point of application in the direction of the force \cite{tipler2007physics}. For instance, a robot arm picks up an object from the floor, and places it on the shelf. The work done on the object is equal to the weight of the object multiplied by the vertical distance to the floor. As a result, the potential energy of the object becomes higher. 

The work-energy principle states that the work done by all forces acting on a particle equals the change in the kinetic energy of the particle \cite{meriam2012engineering}.
That is, the work $W$ done by a force on an object (simplified as a particle) equals the change in the object's kinetic energy $E_{k}$ \cite{young2006sears}, $W=\Delta E_{k}={\frac{1}{2}}mv_{2}^{2}-{\frac{1}{2}}mv_{1}^{2}$, where $v_{1}$ and $v_{2}$ are the speeds of the object before and after the work is done, respectively, and $m$ is its mass. 
As the robot arm is moving the object towards the shelf, the work is being done by the robot on the object.
Consequently, the kinetic energy of the object increases. 


\section{Method}
\label{sec:method}

In this section, we formally describe our approach, including the motivation, the derivation of the trajectory energy function, the energy-based prioritization framework, and a comparison with prioritized experience replay \cite{schaul2015prioritized}. 

\subsection{Motivation}

Consider a robotic manipulation task. We observe that in order to complete the tasks, the robot needs to apply force and do work on the object. 
Typically, the more difficult a task is, the more work from the robot is required. 
Consequently, the energy of the object is changed by the robot. 
Thus, our hypothesis is that, in robotic manipulation tasks, the trajectory energy of the object indicates the difficulty level of the tasks.

From the perspective of curriculum learning, we want to assign the right level of curriculum to the agent. The curriculum should not be too difficult to achieve, also not too simple to learn. We use the trajectory energy to evaluate the difficulty level of the curriculum, and then prioritize the difficult but still achievable tasks for the agent. In this way, the agent might learn with higher sample-efficiency. In robotic tasks, training samples are expensive to acquire, making sample-efficiency in learning important.

\subsection{Trajectory Energy Function}
\label{sec:trajectory_energy_function}

In this section, we introduce the trajectory energy function formally.

A complete trajectory $\mathcal{T}$ in an episode is represented as a tuple $(\mathcal{S}, \mathcal{A}, p, r, \gamma)$. 
A trajectory contains a series of continuous states $s_t$, where $t$ is the timestep $t\in \{0, 1,..,T\}$.
The interval between each timestep, $\Delta t$, corresponds to the sampling frequency in the system, such as $\Delta t = 0.04$ in our experiments.
Each state $s_t \in \mathcal{S}$ also includes the state of the achieved goal, meaning the goal state is a subset of the normal state.
Here, we overwrite the notation $s_t$ as the achieved goal state, i.e.\ the state of the object. 
A trajectory energy function, $E_{traj}$, only depends on the goal states, $s_0, s_1, ..., s_{T}$, and represents the total energy of a trajectory. 

Each state $s_t$ is described as a state vector. In robotic manipulation tasks, we use a state vector $s_t = [x_t, y_t, z_t, a_t, b_t, c_t, d_t]$ to describe the state of the object. 
In each state $s_t$, $x$, $y$, and $z$ specify the object position in the Cartesian coordinate system; $a$, $b$, $c$, and $d$ of a quaternion, $q = a + bi + cj + dk$, describe the orientation of the object. The total trajectory energy consists of three parts, namely the potential energy, the kinetic energy, and the rotational energy.

\textbf{Potential Energy:}
The potential energy of the object $E_p$ at the state $s_t$  is calculated using: 
$
E_p(s_{t})= m g z_{t}
$, 
where $m$ denotes the mass of the object, and $g\approx 9.81~m/s^2$ represents the gravity of earth. 

\textbf{Kinetic Energy:}
To calculate the kinetic energy, we need the velocity of the object. The velocity along the $x$-axis can be calculated using $v_{x,t} \approx (x_{t}-x_{t-1})/\Delta t$. Similarly, the velocities along the $y$-axis and the $z$-axis are calculated as $v_{y,t} \approx (y_{t}-y_{t-1})/\Delta t$ and $v_{z,t} \approx (z_{t}-z_{t-1})/\Delta t$, respectively. The kinetic energy at $s_t$ can be approximated as:
\begin{align*}
E_{k}(s_{t})
=\frac{1}{2} m v^2_{x,t}+\frac{1}{2} m v^2_{y,t}+\frac{1}{2} m v^2_{z,t} 
\approx \frac{m\left((x_{t}-x_{t-1})^2+(y_{t}-y_{t-1})^2+(z_{t}-z_{t-1})^2\right)}{2\Delta t^2}.
\end{align*}

\textbf{Rotational Energy:} For the rotational energy function, we have the quaternion in the simulation \cite{plappert2018multi}, $q = a + bi + cj + dk$, representing the object orientation. First, we need to convert the quaternion representation to Euler angles, $(\phi, \theta, \psi)$, where $\phi$, $\theta$, and $\psi$ represents the rotations around the $x$, $y$, and $z$-axises, respectively. The Euler angles are obtained from quaternion using \cite{blanco2010tutorial}:
\begin{align*}
\left[ \begin{array}{c}
\phi \\
\theta \\
\psi \end{array} \right] =
\left[ \begin{array}{c}
\mathrm{arctan} \frac{2(ab+cd)}{1-2(b^2+c^2)} \\
\mathrm{arcsin} (2(ac - db))\\
\mathrm{arctan} \frac{2(ad+bc)}{1-2(c^2+d^2)} \end{array} \right]=
\left[ \begin{array}{c}
\mathrm{atan2} (2(ab+cd), 1-2(b^2+c^2)) \\
\mathrm{asin} (2(ac-db)) \\
\mathrm{atan2} (2(ad+bc), 1-2(c^2+d^2)) \\
\end{array} \right].
\end{align*}
Note that to obtain full orientations we use $\mathrm{atan2}$ in the implementation instead of the regular $\mathrm{atan}$ function because $\mathrm{atan2}$ allows calculating the arc-tangent of all four quadrants. $\mathrm{Atan}$ only allows calculating of quadrants one and four.
The rotational energy in physics is defined as: $E_r=\frac{1}{2}I\omega^2$, where $I$ is the moment of inertia around the axis of rotation; $\omega$ is the angular velocity and $E_r$ is the rotational energy, also termed as angular kinetic energy. 
The angular velocity around the $x$-axis is: $\omega_{x,t} \approx (\phi_t - \phi_{t-1})/\Delta_t$.
Similarly, for the $y$ and $z$-axises $\omega_{y,t} \approx (\theta_t - \theta_{t-1})/\Delta_t$ and $\omega_{z,t} \approx (\psi_t - \psi_{t-1})/\Delta_t$.
We approximate the rotational energy as:
\begin{align*}
E_r ( s_{t})
= \frac{1}{2} I_x \omega_{x,t}^2 + \frac{1}{2} I_y \omega_{y,t}^2 + \frac{1}{2} I_z \omega_{z,t}^2 
\approx  \frac{ I_x (\phi_t-\phi_{t-1})^2 + I_y (\theta_t-\theta_{t-1})^2 + I_z (\psi_t-\psi_{t-1})^2  }{2\Delta t^2}.
\end{align*}

\textbf{Total Energy:} The total energy is defined as:
$
E (s_{t}) = E_{p} (s_t) + E_{k} (s_t) + E_{r} (s_t).
$
The total energy includes the potential energy, the kinetic energy, and the rotation energy.
Since for prioritizing different trajectories, we are only interested in the relative differences of the trajectory energy, these constant variables, including $m$, $I_x$, $I_y$, and $I_z$, can be set as a constant, such as $m=I_x=I_y=I_z=1$ used in our experiments.

\textbf{Transition Energy:}
We define the transition energy as the total energy increase from the previous state $s_{t-1}$ to the current state $s_t$, mathematically:
\begin{align}
E_{tran}(s_{t-1}, s_{t}) =  \mathrm{clip}\left(E (s_{t}) - E (s_{t-1}), 0, E_{tran}^{max}\right)
\label{eq:transition}
\end{align}
where $t \geq 1$ and $E_{tran}^{max}$ is the predefined maximal transition energy value. The clip function limits the transition energy value in an interval of $[0, E_{tran}^{max}]$. Here, we are only interested in the positive transition energy because the energy increase of the object is only due to the work done by the robot. The robot does work on the object, consequently, the total energy of the object increases.  
In practice, to mitigate the influence of some particular large transition energy, we find it useful to clip the transition energy with a threshold value $E_{tran}^{max}$. This trick makes the training stable. The threshold value can either be tuned as a hyperparameter or estimated using the energy functions.

\textbf{Trajectory Energy:}
Given the definition of the transition energy, we define the trajectory energy as the sum of the transition energy over all the timesteps in the trajectory, mathematically:
\begin{align}
E_{traj}(\mathcal{T}) = E_{traj}(s_0, s_1, ..., s_T) = \sum_{t=1}^{T} E_{tran}(s_{t-1}, s_{t})
\label{eq:trajectory}
\end{align}

\subsection{Energy-Based Prioritization}

In this section, we describe the Energy-Based Prioritization (EBP) framework. In a nutshell, we first calculate the trajectory energy, then prioritize the trajectories with higher energy for replay. 

At the beginning of each episode, the agent uses random policies to start to explore the environment.
The sampled trajectories are stored in a replay buffer. When the agent acquires a new trajectory, the agent calculates the  energy by using the trajectory energy function, Equation (\ref{eq:trajectory}), and stores the energy value along with the trajectory in the replay buffer for later prioritization.

During sampling from the replay buffer, the agent uses the trajectory energy values directly as the probability for sampling. This means that the high energy trajectories have higher priorities to be replayed. Mathematically, the probability of a trajectory $\mathcal{T}_i$ to be replayed after the prioritization is:
\begin{align}
p(\mathcal{T}_i) = \frac{E_{traj}(\mathcal{T}_i)}{\sum_{n=1}^NE_{traj}(\mathcal{T}_n)}
\label{eq:EBP}
\end{align}
where $N$ is the total number of trajectories in the buffer.


\textbf{Complete Algorithm:} We summarize the complete training algorithm in Algorithm~\ref{algo:complete}.

\begin{algorithm}
\caption{HER with Energy-Based Prioritization (EBP)}\label{algo:complete}
\begin{algorithmic}
\State \textbf{Given:}
\State{\hspace{2.5mm} $\bullet$ an off-policy RL algorithm $\mathbb{A}$ \hfill{$\triangleright$ e.g.\ DQN, DDPG}}
\State{\hspace{2.5mm} $\bullet$ a reward function $r : \mathcal{S} \times \mathcal{A} \times \mathcal{G} \rightarrow \mathbb{R}$. \hfill{$\triangleright$ e.g.\ $r(s, a, g) = -1$ if fail, $0$ if success} }
\State{Initialize neural networks of $\mathbb{A}$ and replay buffer $R$}
\For{episode $=$ 1, $N$}
\State{Sample a goal $g$ and an initial state $s_0$.}
\For{$t = 0, T-1$}
\State{Sample an action $a_t$ using the behavioral policy from $\mathbb{A}$:}
\State{\hspace{10mm} $a_t \leftarrow \pi_{b}(s_t\| g)$ \hfill{$\triangleright$ $\|$ denotes concatenation}}
\State{Execute the action $a_t$ and observe a new state $s_{t+1}$}
\EndFor

\State{Calculate trajectory energy $E_{traj}(s_0, s_1, ..., s_T)$ via Equation (\ref{eq:transition}) and (\ref{eq:trajectory}) \hfill{$\triangleright$ trajectory energy}}
\State{Calculate priority $p(\mathcal{T})$ based on Equation (\ref{eq:EBP})}

\For{$t=0, T-1$}
\State{$r_t := r(s_t, a_t, g)$}

\State{Store the transition $(s_t \| g, a_t, r_t, s_{t+1} \| g, p, E_{traj})$ in $R$}
\State{Sample trajectory $\mathcal{T}$ for replay based on priority $p(\mathcal{T})$ \hfill{$\triangleright$ prioritization}}
\State{Sample transitions $(s_t, a_t, s_{t+1})$ from $\mathcal{T}$}
\State{Sample virtual goals $g' \in \{s_{t+1}, ..., s_{T-1}\}$ at a future timestep in $\mathcal{T}$}
\State{$r'_t:= r(s_t, a_t, g')$ \hfill{$\triangleright$ recalculate reward (HER) }}
\State{Store the transition $(s_t\|g', a_t, r'_t, s_{t+1}\|g', p, E_{traj})$ in $R$}
\EndFor
\For{$t=1, M$}
\State{Sample a minibatch $B$ from the replay buffer $R$}
\State{Perform one step of optimization using $\mathbb{A}$ and minibatch $B$}
\EndFor
\EndFor
\end{algorithmic}
\end{algorithm}

\subsection{Comparison with Prioritized Experience Replay}

To the best our knowledge, the most similar method to EBP is Prioritized Experience Replay (PER) \cite{schaul2015prioritized}. 
To combine PER with HER, we calculate the TD-error of each transition based on the randomly selected achieved goals. Then we prioritize the transitions with higher TD-errors for replay. 
It is known that PER can become very expensive in computational time. 
The reason is that PER uses TD-errors for prioritization. After each update of the model, the agent needs to update the priorities of the transitions in the replay buffer with the new TD-errors, and then ranks them based on the priorities and samples the trajectories for replay. In our experiemnts, see Section \ref{sec:experiments}, we use the efficient implementation based on the "sum-tree" data structure, which can be relatively efficiently updated and sampled from \cite{schaul2015prioritized}. 

Compared to PER, EBP is much faster in computational time because it only calculates the trajectory energy once, when a new trajectory becomes available. 
Due to this  reason, EBP is much more efficient than PER in computational time and can easily be combined with any multi-goal RL methods, such as HER.
In the experiments Section \ref{sec:experiments}, we first compare the performance improvement of EBP and PER. 
Afterwards, we compare the time-complexity of PER and EBP. We show that EBP improves performance without additional computational time.
However, PER consumes much more computational time with less improvement.
Furthermore, the motivations of PER and EBP are different. The former uses TD-errors, while the latter is based on the energy in physics.


\section{Experiments}
\label{sec:experiments}

In this section, we first introduce the robot simulation environment used for evaluation. 
Then, we investigate the following questions: \\
\phantom{} - Does incorporating energy-based prioritization bring benefits to hindsight experience replay? \\
\phantom{} - Does energy-based prioritization improve the sample-efficiency in robotic manipulation tasks? \\
\phantom{} - How does the trajectory energy relate to the TD-errors of the trajectory during training?\\
Our code is available at this link\footnote{https://github.com/ruizhaogit/EnergyBasedPrioritization}.

\textbf{Environments:}
The environment we used throughout our experiments is the robotic simulations provided by OpenAI Gym \cite{brockman2016openai,plappert2018multi}, using the MuJoCo physics engine \cite{todorov2012mujoco}. 

\begin{figure}
	\centering
	\includegraphics[width=5. in]{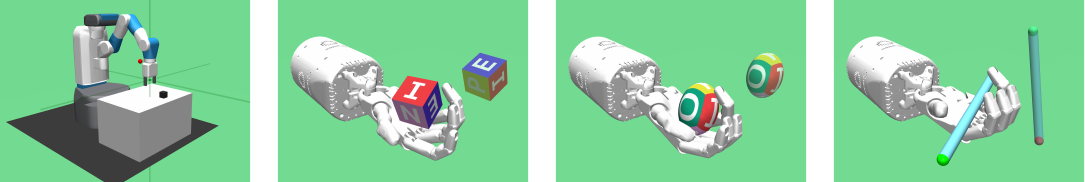}
	\caption{Robot arm Fetch and Shadow Dexterous hand environment: 
	\texttt{FetchPickAndPlace}, \texttt{HandManipulateBlock}, \texttt{HandManipulateEgg}, and \texttt{HandManipulatePen}.}
	\label{fig:fetchhand4env}
\end{figure}

The robotic environment is based on currently existing robotic hardware and is designed as a standard benchmark for Multi-goal RL. The robot agent is required to complete several tasks with different goals in each scenario. There are two kinds of robot agents in the environment. One is a 7-DOF Fetch robotic arm with a two-finger gripper as an end-effector. The other is a 24-DOF Shadow Dexterous robotic hand. We use four challenging tasks for evaluation, including pick \& place, and hand manipulation of block, egg, or pen, see Figure~\ref{fig:fetchhand4env}.

The states in the simulation consist of positions, orientations, linear and angular velocities of all robot joints and of an object. 
Goals represent desired position and orientations of the the object. There is a tolerant range around the desired positions and orientations.
In all environments, we consider sparse rewards. If the object is not in the tolerant range of the goal, the agent receives reward signal -$1$ for each transition; otherwise the reward signal is $0$.

\textbf{Performance:}
To test the performance difference between vanilla HER, HER with PER, and HER with EBP, we run the experiment in all four challenging object manipulation robotic environments. 

\begin{figure}
	\centering
	\includegraphics[width=5.5 in]{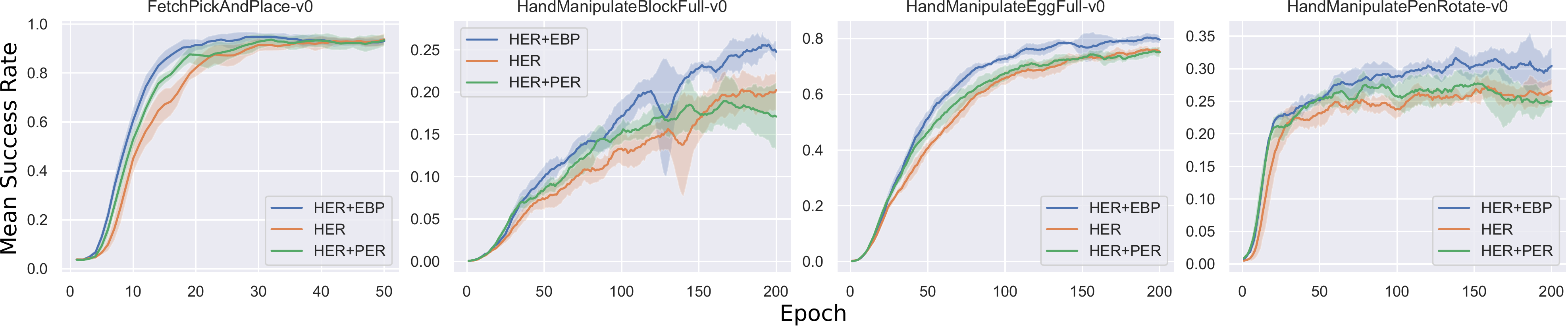}
	\caption{Mean test success rate with standard deviation range for all four environments}
	\label{fig:fig4plot}
\end{figure}

We compare the mean success rates. Each experiment is carried out across 5 random seeds and the shaded
area represents the standard deviation. The learning curve with respect to training epochs is shown in Figure \ref{fig:fig4plot}. For all experiments, we use 19 CPUs. For the robotic arm environment, we train the model for 50 epochs with $E_{tran}^{max}=0.5$ . For the robotic hand environment, we train the agent for 200 epochs with $E_{tran}^{max}=2.5$. After training, we use the best learned policy as the final policy, and test it in the environment. The testing results are the final mean success rates. A comparison of the final performances is shown in Table \ref{tab:results}.

From Figure \ref{fig:fig4plot}, we can see that HER with EBP converges faster than do both vanilla HER and HER with PER in all four tasks. 
The agent trained with EBP also shows a better performance, at the end of the training time.
This is attractive, since HER with EBP consumes nearly the same computational time as vanilla HER, as shown in Table \ref{tab:time}.
However, we see that HER with PER consumes about 10 times the training time as vanilla HER or HER with EBP does in the robot hand environments.

From Table \ref{tab:results}, we can see that HER cooperating with EBP gives a better performance in all four tasks. The improvement varies from 1 percentage point to 5.3 percentage points compared to HER. The average improvement over the four tasks is 3.75 percentage points. 
We can see that EBP is a simple yet effective method, \emph{without increasing computational time}, but still, improves current state-of-the-art methods.

\begin{table}
\centering
\caption{Final Mean Success Rate for all four environments}
\begin{tabular}{ p{2cm} p{2cm} p{2cm} p{2cm} p{2cm} } \hlineB{3}
  					& Pick \& Place 			& Block 						& Egg 						& Pen						\\ 	\hline
Vanilla HER 	& 93.78\%				& 20.32\% 				& 76.19\%	 			& 27.28\%				\\
HER + PER	 	& 93.66\%				& 18.95\% 				& 75.46\%	 			& 27.74\%				\\
HER + EBP 		& \textbf{94.84\%}	& \textbf{25.63\%} 	& \textbf{80.42\%} 	& \textbf{31.69\%}	\\ 	\hlineB{3}
\end{tabular}
\label{tab:results}
\end{table}

\begin{table}
\centering
\caption{Training time (hours:minutes:seconds) in all four environments (single run)}
\begin{tabular}{ p{2cm} p{2cm} p{2cm} p{2cm} p{2cm} } \hlineB{3}
  					& Pick \& Place 			& Block 						& Egg 					& Pen					\\ 	\hline
Vanilla HER 	& 01:32:40				& 08:28:19 				& 07:19:59	 		& 07:33:29			\\
HER + PER 		& 03:07:45				& 80:43:27 				& 79:51:55	 		& 81:10:38			\\
HER + EBP 		& 01:29:57				& 07:28:29 				& 07:28:25 			& 07:35:48			\\ 	\hlineB{3}
\end{tabular}
\label{tab:time}
\end{table}

\begin{figure}
	\centering
	\includegraphics[width=5.5 in]{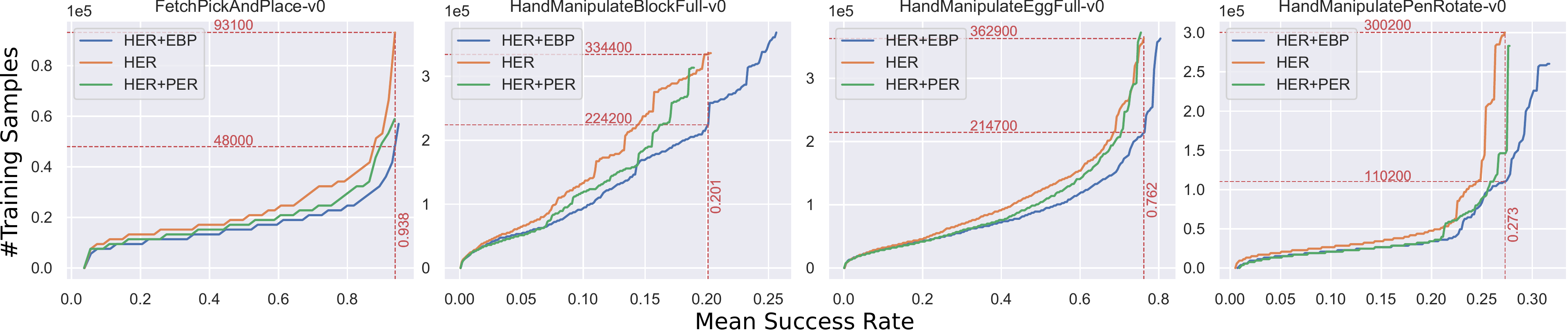}
	\caption{Number of training samples needed with respect to mean test success rate for all four environments (the lower the better)}
	\label{fig:eff4plot}
\end{figure}

\textbf{Sample-Efficiency:}
To compare the sample-efficiency of vanilla HER and HER with EBP, we compare the number of training samples needed for a certain mean test success rate. The comparison is shown in Figure \ref{fig:eff4plot}.

From Figure \ref{fig:eff4plot}, in the \texttt{FetchPickAndPlace-v0} environment, we can see that for the same 93.8\% mean test success rate, HER needs 93,100 samples for training, while HER with EBP only needs 48,000 samples. 
In this case, HER with EBP is nearly twice (1.94) as sample-efficient as vanilla HER. Similarly, in the other three environments, EBP improves sample-efficiency by factors of 1.49, 1.69, and 2.72, respectively. 
In conclusion, for all four testing environments, EBP is able to improve sample-efficiency by an average factor of two (1.96) over vanilla HER's sample-efficiency.

\textbf{Insights:}
\begin{figure}
	\centering
	\includegraphics[width=5.5 in]{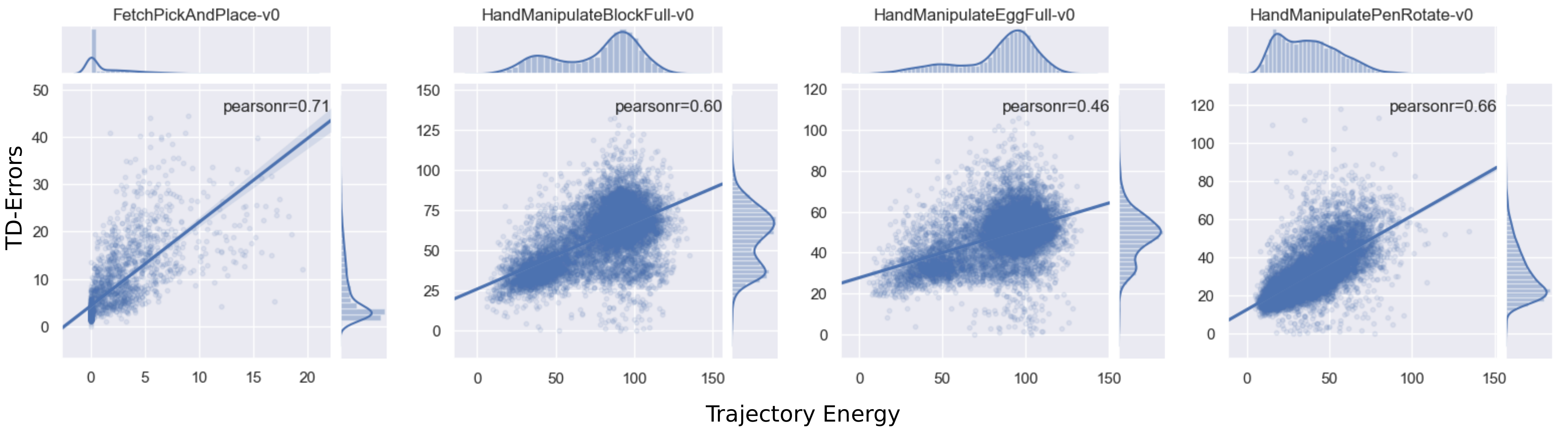}
	\caption{Pearson correlation between the trajectory energy and TD-errors in the middle of training}
	\label{fig:pearson4plot}
\end{figure}
We also investigate the correlation between the trajectory energy and the TD-errors of the trajectory.
The Pearson correlation coefficient, i.e.\ Pearson's r \cite{benesty2009pearson}, between the energy and the TD-errors of the trajectory is shown in Figure \ref{fig:pearson4plot}.
The value of Pearson's r is between 1 and -1, where 1 is total positive linear correlation, 0 is no linear correlation, -1 is total negative linear correlation.
In Figure \ref{fig:pearson4plot}, we can see that the trajectory energy is correlated with the TD-errors of the trajectory with an average Pearson's r of 0.6.
This proves that high energy trajectories are relatively more valuable for learning. Therefore, it is helpful to prioritize high energy trajectories during training. 
 

\section{Related Work}
\label{sec:related_work}

Experience replay was proposed by \citet{lin1992self} in 1992 and became popular due to the success of DQN \cite{mnih2015human} in 2015. In the same year, prioritized experience replay was introduced by \citet{schaul2015prioritized} as an improvement of the experience replay in DQN. It prioritized the transitions with higher TD-error in the replay buffer to speed up training. This idea is complementary to our method. 

In 2015, \citet{schaul2015universal} proposed universal function approximators, generalizing not just over states but also over goals. 
There are also many other research works about multi-task RL \cite{schmidhuber1990learning, caruana1998multitask, da2012learning, kober2012reinforcement, pinto2017learning, foster2002structure, sutton2011horde}. 
Hindsight experience replay \cite{andrychowicz2017hindsight} is a kind of goal-conditioned RL that substitutes any achieved goals as real goals to encourage the agent to learn something instead of nothing.

Our method can be considered as a form of curriculum learning \cite{elman1993learning,bengio2009curriculum,zaremba2014learning, graves2017automated,sukhbaatar2017intrinsic,srivastava2013first}. The essence of our method is to assign priority to the achieved trajectories with higher energy, which are relatively difficult to achieve. In RL,  curriculum generation can be traced back to 2004. \citet{schmidhuber2004optimal} used a program search to construct an asymptotically optimal algorithm to approach the problem. 
Recently, \citet{florensa2017reverse} trained the agent reversely, from the start states near the goal states, gradually to the states far from the goals.
Our method bares a similar motivation, but is orthogonal to these previous works and can be combined with them.       


\section{Conclusion}

In conclusion, we proposed a simple yet effective energy-based approach to prioritize hindsight experience. 
Energy-Based Prioritization shows promising experimental results in all four challenging robotic manipulation tasks.
This method can be combined with any off-policy RL methods. 
We integrated physics knowledge about energy into the modern reinforcement learning paradigm, and improved sample-efficiency by a factor of two and the final performance by about four percentage points on top of state-of-the-art methods, without increasing computational time.



\clearpage


\bibliography{reference}

\begin{thebibliography}{46}
\providecommand{\natexlab}[1]{#1}
\providecommand{\url}[1]{\texttt{#1}}
\expandafter\ifx\csname urlstyle\endcsname\relax
  \providecommand{\doi}[1]{doi: #1}\else
  \providecommand{\doi}{doi: \begingroup \urlstyle{rm}\Url}\fi

\bibitem[Sutton and Barto(1998)]{sutton1998reinforcement}
R.~S. Sutton and A.~G. Barto.
\newblock \emph{Reinforcement learning: An introduction}, volume~1.
\newblock MIT press Cambridge, 1998.

\bibitem[Goodfellow et~al.(2016)Goodfellow, Bengio, Courville, and
  Bengio]{goodfellow2016deep}
I.~Goodfellow, Y.~Bengio, A.~Courville, and Y.~Bengio.
\newblock \emph{Deep learning}, volume~1.
\newblock MIT press Cambridge, 2016.

\bibitem[Zhao et~al.(2017)Zhao, Ali, and Van~der Smagt]{zhao2017two}
R.~Zhao, H.~Ali, and P.~Van~der Smagt.
\newblock Two-stream rnn/cnn for action recognition in 3d videos.
\newblock In \emph{2017 IEEE/RSJ International Conference on Intelligent Robots
  and Systems (IROS)}, pages 4260--4267. IEEE, 2017.

\bibitem[Mnih et~al.(2015)Mnih, Kavukcuoglu, Silver, Rusu, Veness, Bellemare,
  Graves, Riedmiller, Fidjeland, Ostrovski, et~al.]{mnih2015human}
V.~Mnih, K.~Kavukcuoglu, D.~Silver, A.~A. Rusu, J.~Veness, M.~G. Bellemare,
  A.~Graves, M.~Riedmiller, A.~K. Fidjeland, G.~Ostrovski, et~al.
\newblock Human-level control through deep reinforcement learning.
\newblock \emph{Nature}, 518\penalty0 (7540):\penalty0 529, 2015.

\bibitem[Silver et~al.(2016)Silver, Huang, Maddison, Guez, Sifre, Van
  Den~Driessche, Schrittwieser, Antonoglou, Panneershelvam, Lanctot,
  et~al.]{silver2016mastering}
D.~Silver, A.~Huang, C.~J. Maddison, A.~Guez, L.~Sifre, G.~Van Den~Driessche,
  J.~Schrittwieser, I.~Antonoglou, V.~Panneershelvam, M.~Lanctot, et~al.
\newblock Mastering the game of go with deep neural networks and tree search.
\newblock \emph{nature}, 529\penalty0 (7587):\penalty0 484–489, 2016.

\bibitem[Bordes et~al.(2016)Bordes, Boureau, and Weston]{bordes2016learning}
A.~Bordes, Y.-L. Boureau, and J.~Weston.
\newblock Learning end-to-end goal-oriented dialog.
\newblock \emph{arXiv preprint arXiv:1605.07683}, 2016.

\bibitem[Zhao and Tresp(2018{\natexlab{a}})]{zhao2018improving}
R.~Zhao and V.~Tresp.
\newblock Improving goal-oriented visual dialog agents via advanced recurrent
  nets with tempered policy gradient.
\newblock \emph{arXiv preprint arXiv:1807.00737}, 2018{\natexlab{a}}.

\bibitem[Zhao and Tresp(2018{\natexlab{b}})]{zhao2018learning}
R.~Zhao and V.~Tresp.
\newblock Learning goal-oriented visual dialog via tempered policy gradient.
\newblock In \emph{2018 IEEE Spoken Language Technology Workshop (SLT)}, pages
  868--875. IEEE, 2018{\natexlab{b}}.

\bibitem[Zhao and Tresp(2018{\natexlab{c}})]{zhao2018efficient}
R.~Zhao and V.~Tresp.
\newblock Efficient dialog policy learning via positive memory retention.
\newblock In \emph{2018 IEEE Spoken Language Technology Workshop (SLT)}, pages
  823--830. IEEE, 2018{\natexlab{c}}.

\bibitem[Ng et~al.(2006)Ng, Coates, Diel, Ganapathi, Schulte, Tse, Berger, and
  Liang]{ng2006autonomous}
A.~Y. Ng, A.~Coates, M.~Diel, V.~Ganapathi, J.~Schulte, B.~Tse, E.~Berger, and
  E.~Liang.
\newblock Autonomous inverted helicopter flight via reinforcement learning.
\newblock In \emph{Experimental Robotics IX}, page 363–372. Springer, 2006.

\bibitem[Peters and Schaal(2008)]{peters2008reinforcement}
J.~Peters and S.~Schaal.
\newblock Reinforcement learning of motor skills with policy gradients.
\newblock \emph{Neural networks}, 21\penalty0 (4):\penalty0 682--697, 2008.

\bibitem[Levine et~al.(2016)Levine, Finn, Darrell, and Abbeel]{levine2016end}
S.~Levine, C.~Finn, T.~Darrell, and P.~Abbeel.
\newblock End-to-end training of deep visuomotor policies.
\newblock \emph{The Journal of Machine Learning Research}, 17\penalty0
  (1):\penalty0 1334–1373, 2016.

\bibitem[Chebotar et~al.(2017)Chebotar, Kalakrishnan, Yahya, Li, Schaal, and
  Levine]{chebotar2017path}
Y.~Chebotar, M.~Kalakrishnan, A.~Yahya, A.~Li, S.~Schaal, and S.~Levine.
\newblock Path integral guided policy search.
\newblock In \emph{Robotics and Automation (ICRA), 2017 IEEE International
  Conference on}, pages 3381--3388. IEEE, 2017.

\bibitem[Andrychowicz et~al.(2017)Andrychowicz, Wolski, Ray, Schneider, Fong,
  Welinder, McGrew, Tobin, Abbeel, and Zaremba]{andrychowicz2017hindsight}
M.~Andrychowicz, F.~Wolski, A.~Ray, J.~Schneider, R.~Fong, P.~Welinder,
  B.~McGrew, J.~Tobin, O.~P. Abbeel, and W.~Zaremba.
\newblock Hindsight experience replay.
\newblock In \emph{Advances in Neural Information Processing Systems}, page
  5048–5058, 2017.

\bibitem[Schaul et~al.(2015)Schaul, Horgan, Gregor, and
  Silver]{schaul2015universal}
T.~Schaul, D.~Horgan, K.~Gregor, and D.~Silver.
\newblock Universal value function approximators.
\newblock In \emph{International Conference on Machine Learning}, pages
  1312--1320, 2015.

\bibitem[Schmidhuber(2013)]{schmidhuber2013powerplay}
J.~Schmidhuber.
\newblock Powerplay: Training an increasingly general problem solver by
  continually searching for the simplest still unsolvable problem.
\newblock \emph{Frontiers in psychology}, 4:\penalty0 313, 2013.

\bibitem[Deisenroth et~al.(2014)Deisenroth, Englert, Peters, and
  Fox]{deisenroth2014multi}
M.~P. Deisenroth, P.~Englert, J.~Peters, and D.~Fox.
\newblock Multi-task policy search for robotics.
\newblock In \emph{Robotics and Automation (ICRA), 2014 IEEE International
  Conference on}, pages 3876--3881. IEEE, 2014.

\bibitem[Zhu et~al.(2017)Zhu, Mottaghi, Kolve, Lim, Gupta, Fei-Fei, and
  Farhadi]{zhu2017target}
Y.~Zhu, R.~Mottaghi, E.~Kolve, J.~J. Lim, A.~Gupta, L.~Fei-Fei, and A.~Farhadi.
\newblock Target-driven visual navigation in indoor scenes using deep
  reinforcement learning.
\newblock In \emph{2017 IEEE international conference on robotics and
  automation (ICRA)}, pages 3357--3364. IEEE, 2017.

\bibitem[Held et~al.(2017)Held, Geng, Florensa, and Abbeel]{held2017automatic}
D.~Held, X.~Geng, C.~Florensa, and P.~Abbeel.
\newblock Automatic goal generation for reinforcement learning agents.
\newblock \emph{arXiv preprint arXiv:1705.06366}, 2017.

\bibitem[Plappert et~al.(2018)Plappert, Andrychowicz, Ray, McGrew, Baker,
  Powell, Schneider, Tobin, Chociej, Welinder, et~al.]{plappert2018multi}
M.~Plappert, M.~Andrychowicz, A.~Ray, B.~McGrew, B.~Baker, G.~Powell,
  J.~Schneider, J.~Tobin, M.~Chociej, P.~Welinder, et~al.
\newblock Multi-goal reinforcement learning: Challenging robotics environments
  and request for research.
\newblock \emph{arXiv preprint arXiv:1802.09464}, 2018.

\bibitem[Todorov et~al.(2012)Todorov, Erez, and Tassa]{todorov2012mujoco}
E.~Todorov, T.~Erez, and Y.~Tassa.
\newblock Mujoco: A physics engine for model-based control.
\newblock In \emph{Intelligent Robots and Systems (IROS), 2012 IEEE/RSJ
  International Conference on}, pages 5026--5033. IEEE, 2012.

\bibitem[Brockman et~al.(2016)Brockman, Cheung, Pettersson, Schneider,
  Schulman, Tang, and Zaremba]{brockman2016openai}
G.~Brockman, V.~Cheung, L.~Pettersson, J.~Schneider, J.~Schulman, J.~Tang, and
  W.~Zaremba.
\newblock Openai gym.
\newblock \emph{arXiv preprint arXiv:1606.01540}, 2016.

\bibitem[Lillicrap et~al.(2016)Lillicrap, Hunt, Pritzel, Heess, Erez, Tassa,
  Silver, and Wierstra]{lillicrap2015continuous}
T.~P. Lillicrap, J.~J. Hunt, A.~Pritzel, N.~Heess, T.~Erez, Y.~Tassa,
  D.~Silver, and D.~Wierstra.
\newblock Continuous control with deep reinforcement learning.
\newblock In \emph{International Conference on Learning Representations}, 2016.

\bibitem[Polyak and Juditsky(1992)]{polyak1992acceleration}
B.~T. Polyak and A.~B. Juditsky.
\newblock Acceleration of stochastic approximation by averaging.
\newblock \emph{SIAM Journal on Control and Optimization}, 30\penalty0
  (4):\penalty0 838--855, 1992.

\bibitem[Tipler and Mosca(2007)]{tipler2007physics}
P.~A. Tipler and G.~Mosca.
\newblock \emph{Physics for scientists and engineers}.
\newblock Macmillan, 2007.

\bibitem[Meriam and Kraige(2012)]{meriam2012engineering}
J.~L. Meriam and L.~G. Kraige.
\newblock \emph{Engineering mechanics: dynamics}, volume~2.
\newblock John Wiley \& Sons, 2012.

\bibitem[Young et~al.(2006)Young, Freedman, and Ford]{young2006sears}
H.~D. Young, R.~A. Freedman, and A.~L. Ford.
\newblock \emph{Sears and Zemansky's university physics}, volume~1.
\newblock Pearson education, 2006.

\bibitem[Schaul et~al.(2016)Schaul, Quan, Antonoglou, and
  Silver]{schaul2015prioritized}
T.~Schaul, J.~Quan, I.~Antonoglou, and D.~Silver.
\newblock Prioritized experience replay.
\newblock In \emph{International Conference on Learning Representations}, 2016.

\bibitem[Blanco(2010)]{blanco2010tutorial}
J.-L. Blanco.
\newblock A tutorial on se (3) transformation parameterizations and on-manifold
  optimization.
\newblock \emph{University of Malaga, Tech. Rep}, 3, 2010.

\bibitem[Benesty et~al.(2009)Benesty, Chen, Huang, and
  Cohen]{benesty2009pearson}
J.~Benesty, J.~Chen, Y.~Huang, and I.~Cohen.
\newblock Pearson correlation coefficient.
\newblock In \emph{Noise reduction in speech processing}, pages 1--4. Springer,
  2009.

\bibitem[Lin(1992)]{lin1992self}
L.-J. Lin.
\newblock Self-improving reactive agents based on reinforcement learning,
  planning and teaching.
\newblock \emph{Machine learning}, 8\penalty0 (3-4):\penalty0 293–321, 1992.

\bibitem[Schmidhuber and Huber(1990)]{schmidhuber1990learning}
J.~Schmidhuber and R.~Huber.
\newblock \emph{Learning to generate focus trajectories for attentive vision}.
\newblock Institut f{\"u}r Informatik, 1990.

\bibitem[Caruana(1998)]{caruana1998multitask}
R.~Caruana.
\newblock Multitask learning.
\newblock In \emph{Learning to learn}, page 95–133. Springer, 1998.

\bibitem[Da~Silva et~al.(2012)Da~Silva, Konidaris, and Barto]{da2012learning}
B.~Da~Silva, G.~Konidaris, and A.~Barto.
\newblock Learning parameterized skills.
\newblock \emph{arXiv preprint arXiv:1206.6398}, 2012.

\bibitem[Kober et~al.(2012)Kober, Wilhelm, Oztop, and
  Peters]{kober2012reinforcement}
J.~Kober, A.~Wilhelm, E.~Oztop, and J.~Peters.
\newblock Reinforcement learning to adjust parametrized motor primitives to new
  situations.
\newblock \emph{Autonomous Robots}, 33\penalty0 (4):\penalty0 361--379, 2012.

\bibitem[Pinto and Gupta(2017)]{pinto2017learning}
L.~Pinto and A.~Gupta.
\newblock Learning to push by grasping: Using multiple tasks for effective
  learning.
\newblock In \emph{Robotics and Automation (ICRA), 2017 IEEE International
  Conference on}, pages 2161--2168. IEEE, 2017.

\bibitem[Foster and Dayan(2002)]{foster2002structure}
D.~Foster and P.~Dayan.
\newblock Structure in the space of value functions.
\newblock \emph{Machine Learning}, 49\penalty0 (2-3):\penalty0 325--346, 2002.

\bibitem[Sutton et~al.(2011)Sutton, Modayil, Delp, Degris, Pilarski, White, and
  Precup]{sutton2011horde}
R.~S. Sutton, J.~Modayil, M.~Delp, T.~Degris, P.~M. Pilarski, A.~White, and
  D.~Precup.
\newblock Horde: A scalable real-time architecture for learning knowledge from
  unsupervised sensorimotor interaction.
\newblock In \emph{The 10th International Conference on Autonomous Agents and
  Multiagent Systems-Volume 2}, pages 761--768. International Foundation for
  Autonomous Agents and Multiagent Systems, 2011.

\bibitem[Elman(1993)]{elman1993learning}
J.~L. Elman.
\newblock Learning and development in neural networks: The importance of
  starting small.
\newblock \emph{Cognition}, 48\penalty0 (1):\penalty0 71--99, 1993.

\bibitem[Bengio et~al.(2009)Bengio, Louradour, Collobert, and
  Weston]{bengio2009curriculum}
Y.~Bengio, J.~Louradour, R.~Collobert, and J.~Weston.
\newblock Curriculum learning.
\newblock In \emph{Proceedings of the 26th annual international conference on
  machine learning}, pages 41--48. ACM, 2009.

\bibitem[Zaremba and Sutskever(2014)]{zaremba2014learning}
W.~Zaremba and I.~Sutskever.
\newblock Learning to execute.
\newblock \emph{arXiv preprint arXiv:1410.4615}, 2014.

\bibitem[Graves et~al.(2017)Graves, Bellemare, Menick, Munos, and
  Kavukcuoglu]{graves2017automated}
A.~Graves, M.~G. Bellemare, J.~Menick, R.~Munos, and K.~Kavukcuoglu.
\newblock Automated curriculum learning for neural networks.
\newblock \emph{arXiv preprint arXiv:1704.03003}, 2017.

\bibitem[Sukhbaatar et~al.(2017)Sukhbaatar, Lin, Kostrikov, Synnaeve, Szlam,
  and Fergus]{sukhbaatar2017intrinsic}
S.~Sukhbaatar, Z.~Lin, I.~Kostrikov, G.~Synnaeve, A.~Szlam, and R.~Fergus.
\newblock Intrinsic motivation and automatic curricula via asymmetric
  self-play.
\newblock \emph{arXiv preprint arXiv:1703.05407}, 2017.

\bibitem[Srivastava et~al.(2013)Srivastava, Steunebrink, and
  Schmidhuber]{srivastava2013first}
R.~K. Srivastava, B.~R. Steunebrink, and J.~Schmidhuber.
\newblock First experiments with powerplay.
\newblock \emph{Neural Networks}, 41:\penalty0 130--136, 2013.

\bibitem[Schmidhuber(2004)]{schmidhuber2004optimal}
J.~Schmidhuber.
\newblock Optimal ordered problem solver.
\newblock \emph{Machine Learning}, 54\penalty0 (3):\penalty0 211--254, 2004.

\bibitem[Florensa et~al.(2017)Florensa, Held, Wulfmeier, and
  Abbeel]{florensa2017reverse}
C.~Florensa, D.~Held, M.~Wulfmeier, and P.~Abbeel.
\newblock Reverse curriculum generation for reinforcement learning.
\newblock \emph{arXiv preprint arXiv:1707.05300}, 2017.

\end{thebibliography}

\end{document}